# Knowledge and Reasoning Augmented Learning for Language Model Assisted Antimicrobial Decision Support

Zhe Li[1,#], Yehan Qiu[2,#], Yuntian Huang[4,#], Xin Ding[2], Yujie Chen[2], Jing Sun[2], Min Yan[4], Lei Chen[4], Zheng Li[4], Yanfei Shen[4], Mengxi Qie[4], Xin Meng[4], Jiahui Yin[4], Yanwen Fu[2], Yaguang Li[1], Feng Zhang[1], Xinping Zhang[1], Xiang Zhou[1,2,*]

1 Information Center, State Key Laboratory of Complex Severe and Rare Diseases, Peking Union Medical College Hospital, Beijing 100730, China

2 Department of Critical Care Medicine, State Key Laboratory of Complex Severe and Rare Diseases, Peking Union Medical College Hospital, Peking Union Medical College and Chinese Academy of Medical Sciences, Beijing 100730, China

3 Medical Intensive Care Unit, State Key Laboratory of Complex Severe and Rare Diseases, Peking Union Medical College Hospital, Peking Union Medical College and Chinese Academy of Medical Sciences, Beijing 100730, China

4 Beijing Data Pilot Zone Service Co., Ltd., Beijing 101111, China

*# These authors contributed equally: Zhe Li, Yehan Qiu, Yuntian Huang. * Corresponding author: Xiang Zhou.*

## Abstract

**Background:** Antimicrobial decision support requires both clinical knowledge and patient-specific reasoning. We developed Knowledge and Reasoning Augmented Learning (KRAL) to improve language model performance on antimicrobial decision-making tasks.

**Methods:** In this retrospective technical study, a 27-billion-parameter student model underwent guideline-grounded supervised fine-tuning followed by Multi-turn Agentic GRPO using clinical reasoning trajectories. Six configurations were evaluated on 123 clinician-validated clinical antimicrobial treatment cases from Peking Union Medical College Hospital (PUMCH) and 100 JAMA Clinical Challenge cases. A language model judge scored responses, and paired tests compared configurations. Component ablations assessed benchmark performance and training stability.

**Results:** On the PUMCH benchmark, KRAL achieved the highest composite score (0.895 on a 0–1 scale), outperforming all five controls (all $P < 0.001$). On JAMA, KRAL scored 0.600 and exceeded the base model with and without retrieval ($P = 0.0048$ and 0.0499, respectively). Reinforcement learning ablation experiments demonstrate that the reward smoothing can significantly improve the performance and training stability of vanilla GRPO.

**Conclusions:** KRAL improved performance on a single-center clinical antimicrobial benchmark and publicly available JAMA clinical challenge cases by combining guideline-grounded learning with Multi-turn Agentic GRPO. Further external clinical validation is needed to evaluate the practical utility and effectiveness of KRAL.

## Introduction

Antimicrobial agents are widely used for treatment and prophylaxis in outpatient and inpatient care. Optimizing their use remains a persistent clinical challenge: insufficient treatment may allow infection to progress, whereas inappropriate or unnecessarily broad-spectrum therapy may cause toxicity and accelerate antimicrobial resistance. Antimicrobial stewardship therefore requires effective treatment while limiting avoidable drug exposure and harm [1].

Antimicrobial prescribing requires clinical reasoning under uncertainty. Decisions depend on infection site and severity, likely pathogens, local resistance patterns, host immune status and organ function, as well as antimicrobial spectrum, tissue penetration, and pharmacokinetic/pharmacodynamic properties.

Clinicians must reconcile recommendations from current guidelines with incomplete or evolving patient information, often under time constraints, and revise treatment as new evidence emerges. The availability of guidelines does not by itself ensure timely, individualized drug selection, dosing, or treatment adjustment. Decision support must therefore connect applicable knowledge with the circumstances of the individual patient.

Large language models (LLMs) can interpret clinical narratives and answer medical questions [2, 3]. Antimicrobial decision-support studies have evaluated free-text recommendations [4], compared model recommendations with clinician prescribing [5], and used retrieval-augmented generation (RAG) to provide advice grounded in institutional guidelines [6]. External model services also raise privacy and governance concerns [7–9]. Biomedical language models have supported text representation and information extraction [10–12], while specialized models have addressed personalized antibiograms, resistance prediction, intravenous-to-oral switching, and treatment-effect subgroups [13–16]. These developments support access to knowledge and individual prescribing tasks, but evaluating the quality of a generated recommendation does not establish how to train a model to use guideline evidence within complex case reasoning. Access to retrieved guidance alone does not teach a model to select and apply the evidence needed for a specific case.

We developed Knowledge and Reasoning Augmented Learning (KRAL) to address these two requirements in a locally deployable 27-billion-parameter student model (Figure 1). Retrieval-grounded teacher distillation and warmup supervised fine-tuning (SFT) target knowledge alignment using institutional antimicrobial guidance. Multi-turn Agentic Group Relative Policy Optimization (GRPO) uses teacher-generated reasoning–retrieval trajectories from inpatient records and rewards for retrieval actions and clinical responses to target reasoning alignment. Guideline retrieval also provides access to evidence during training and, when enabled, inference. We compared configurations on clinician-validated clinical antimicrobial treatment cases and public clinical vignettes to examine performance and the contribution of inference-time retrieval. Component ablations assessed training stability separately.

# Methods

## Study design and ethics

This retrospective technical study evaluated model development and performance using institutional clinical data and public vignettes. The Peking Union Medical College Hospital (PUMCH) Ethics Committee approved retrospective use of de-identified electronic medical records (I-23PJ1416) and waived informed consent. Institutional guidance, clinical decision support system (CDSS) records, inpatient records, and benchmark data remain subject to privacy, governance, and licensing restrictions.

## Data sources and partitions

Training sources comprised 750 pages of PUMCH antimicrobial guidance, 84 random CDSS questions, and 710 de-identified inpatient records (Table 1). Optical character recognition and segmentation produced 4,157 guideline chunks, and CDSS cleaning retained 79 questions. After distillation and deduplication by normalized full-question text, the chain-of-thought (CoT) dataset was split into 35,122 training and 1,849 validation records using seed 42. Warmup SFT used 35,122 CoT training records and included no JAMA cases. The 710 inpatient records were selected from 2024 admissions with an antimicrobial order and an infection-related diagnosis, including pneumonia, enteritis, typhoid fever, dysentery, tuberculosis, or viral encephalitis. One admission was retained per patient; readmissions and records lacking the required clinical context or a documented target recommendation were excluded. Antimicrobial names were normalized from the order text before matching.

Evaluation used 123 clinician-validated real-world clinical antimicrobial treatment cases from PUMCH and 100 selected English JAMA Clinical Challenge cases [17]. The JAMA answer options were removed; tasks requested a diagnosis or next clinical step and extended beyond antimicrobial prescribing. Models received the case text and, under RAG conditions, retrieved evidence. Reference answers and explanatory scoring criteria were supplied only to the evaluation judge. The separate 200-case JAMA collection and the 100-case evaluation set were deduplicated case by case and shared no cases. The 710 inpatient records used for trajectory generation had no patient-level or record-level overlap with the 123 PUMCH evaluation cases.

## Framework design and comparisons

KRAL linked guideline-grounded distillation, warmup SFT, and Multi-turn Agentic GRPO. Retrieved evidence served two roles: it supported teacher supervision and GRPO training, and it could be supplied again during benchmark inference. The evaluation varied inference-time retrieval separately from the training configuration.

Framework comparisons crossed three training levels with the presence or absence of inference-time RAG: BASE, BASE+RAG, SFT, SFT+RAG, GRPO, and KRAL (full GRPO+RAG). All used the Qwen3.8-27B backbone. In the figures, “KRAL w/o Stage1” denotes full GRPO without distillation and inference-time RAG, and “KRAL w/o Stage2” denotes SFT+RAG without GRPO. All GRPO configurations retained SFT initialization. The framework comparisons therefore examined inference-time retrieval and the addition of GRPO.

## Evidence retrieval and teacher distillation

DeepSeek-V4.1-Flash served as the teacher and Qwen3.8-27B as the student. The shared antimicrobial guideline corpus supported BGE-M3 hybrid retrieval, combining dense, sparse, and ColBERT scores with weights of 0.40, 0.20, and 0.40 and a threshold of 0.3. Answer distillation used the top five chunks. Teacher trajectory generation returned up to three chunks per retrieval term, and evaluation RAG used the top three chunks.

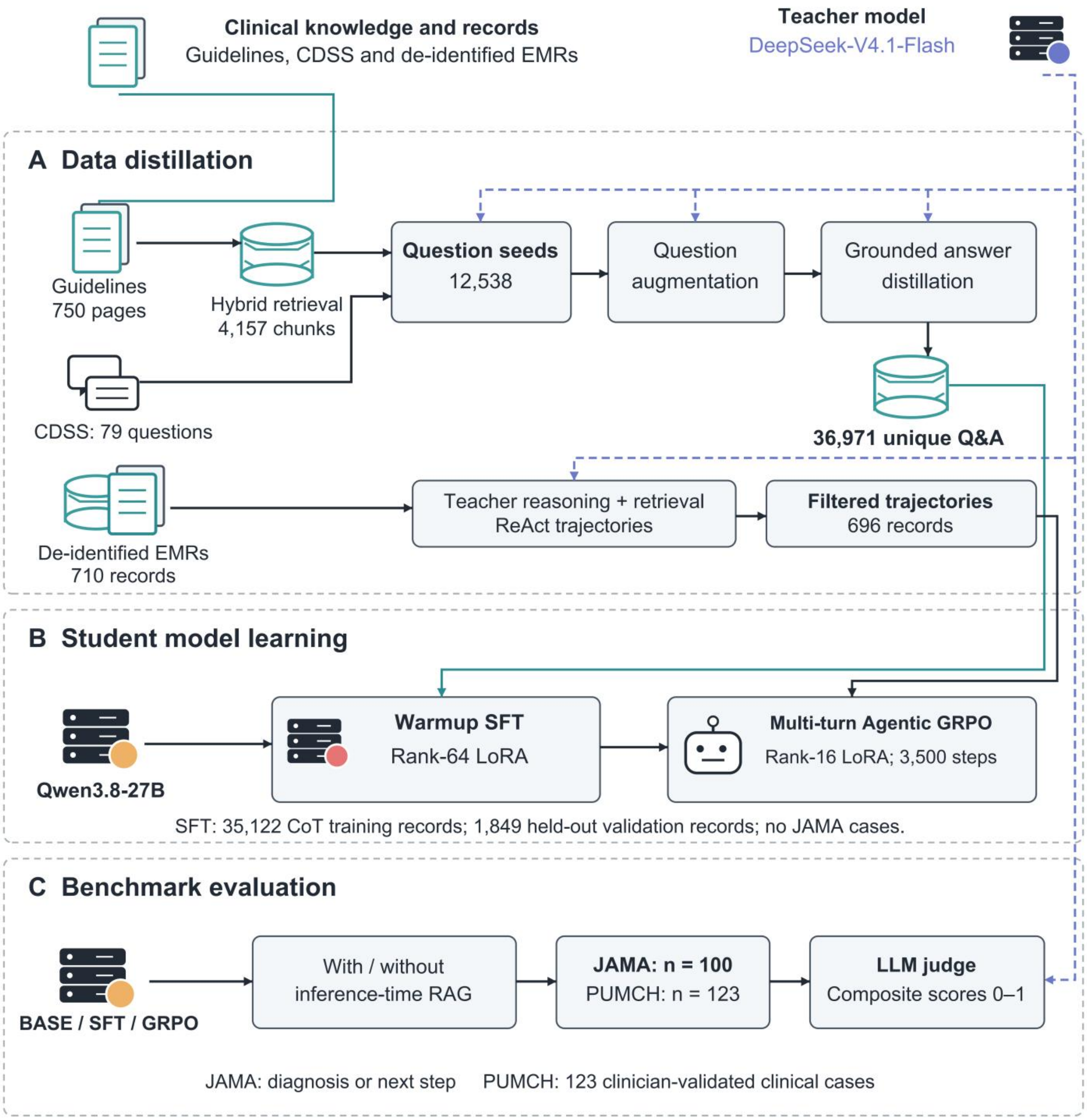

*Figure 1. KRAL workflow for teacher-guided data distillation, student model learning, and clinical benchmark evaluation. Dashed blue arrows denote teacher generation and judging; solid arrows denote data or model flow. CDSS, clinical decision support system; EMRs, electronic medical records; Q&A, question–answer; SFT, supervised fine-tuning; GRPO, Group Relative Policy Optimization; RAG, retrieval-augmented generation.*

The teacher generated three complementary questions per usable guideline chunk, yielding 12,459 questions from 4,153 chunks. Combining these with 79 cleaned CDSS questions produced 12,538 seeds. Two additional variants per seed expanded the collection to 37,614 questions. Retrieval-grounded answer generation and removal of empty reasoning or answers yielded 37,117 CoT records; removing 146 duplicate questions left 36,971 unique records. The final SFT examples paired questions with reasoning and answers. Retrieved evidence was available to the teacher but omitted from these SFT examples.

For GRPO supervision, the teacher independently processed each of the 710 inpatient records through a ReAct-style reasoning–retrieval loop [18]. The loop allowed up to ten retrieval rounds and could stop earlier when evidence was deemed sufficient. Prompt normalization, assistant-format correction, and teacher-based relevance filtering retained 696 trajectories, which provided reference retrieval actions and clinical responses for reward computation. Generated reasoning served as a training signal for the student model.

## Warmup supervised fine-tuning

Warmup SFT adapted Qwen3.8-27B to the distilled institutional question-answering tasks using 35,122 CoT training records, with 1,849 separate CoT records held out for validation (Table 2). Each example paired a question with teacher-generated reasoning and a final answer, training the student to generate reasoning and a response conditioned on the question. Retrieved passages used during teacher distillation were omitted from the student input. This stage provided a task-specific initialization for subsequent Multi-turn Agentic GRPO. Low-rank adaptation (LoRA) used a rank of 64, a scaling parameter alpha of 128, and dropout of 0.05 [19].

Training followed a warmup schedule with an initial learning rate of $2 \times 10^{-4}$, a warmup fraction of 0.05, and three planned epochs. The microbatch size was 2 per GPU with four gradient-accumulation steps; weight decay was 0.1, the gradient-clipping threshold was 1.0, and the NEFTune parameter was 5. Eight H200 GPUs supported bfloat16 training with DeepSpeed ZeRO-3 [20]. Sequence packing improved token utilization, while gradient checkpointing reduced activation-memory requirements. The maximum sequence length was 8,192 tokens. Held-out CoT validation records did not receive gradient updates.

## Multi-turn Agentic GRPO

Multi-turn Agentic GRPO initialized from the SFT checkpoint and used the 696 filtered trajectories. Four H200 GPUs trained the policy and four supported vLLM rollouts. Each case generated eight responses with at most four retrieval rounds. Group-normalized rewards avoided a separate learned value network [21, 22]. Training used rank-16 LoRA, ZeRO-2, bfloat16, a learning rate of $5 \times 10{-}6$, a Kullback–Leibler (KL) coefficient of 0.001, and a 3,500-step budget. The best checkpoint maximized the rolling mean reward over 20 steps.

The reward combined TraceAccuracy and RepetitionPenalty with weights of 1.0 and 0.1. TraceAccuracy evaluated retrieval actions using character/subword Jaccard overlap and clinical responses using reward smoothing (continuous hybrid similarity matching). ColBERT late interaction [23] and BGE-M3 representations [24] supported semantic matching. Probability-ratio clipping used epsilon = 0.1 and epsilon_high = 0.4.

Four matched component ablations removed the repetition penalty, reduced epsilon_high to 0.1, disabled reward smoothing, or replaced character/subword Jaccard with jieba word-set Jaccard. Thus, the subword ablation changed the matching unit rather than introducing whole-term exact matching. Component ablations were evaluated separately from the framework comparisons, with and without inference-time RAG.

Training stability was assessed using gradient norms, valid-budget coverage, and failure-penalized reward area under the curve (AUC). Reward AUC was the failure-penalized mean reward over the fixed 3,500-step budget, with invalid rewards and post-failure steps assigned zero. Coverage was the proportion of the budget contributing valid rewards. These training summaries were assessed separately from benchmark performance. Repeated-seed uncertainty estimates were not available.

### Benchmark scoring and inference settings

We used the established LLM-as-judge approach, in which a capable language model evaluates generated responses against reference answers and explicit scoring criteria [25]. DeepSeek-V4.1-Flash rated each of five dimensions from 0 to 10:

a. Clinical correctness, accepting synonyms, equivalent or more specific diagnoses, and reasonable broader formulations when the core conclusion agreed with the reference;

b. Clinical reasoning quality, rewarding identification of decisive clues even in brief reasoning;

c. Fidelity to the case, penalizing invented findings or histories and unjustified certainty while allowing reasonable inference;

d. Completeness, awarding partial credit for a correct but incomplete core answer;

e. Relevance and concision, rewarding direct answers without unrelated material or a requirement for lengthy explanations.

Case-level benchmark scores were reported on a normalized 0–1 scale. The PUMCH overall score was the arithmetic mean of case-level scores across all 123 cases. Scores reflected graded response quality rather than accuracy or Pass@1 and were interpreted within each benchmark. Generation used temperature 0.1, seed 42, and thinking enabled. Every model was allowed 8,192 generated tokens and a 6,144-token thinking budget. The reported SEM summarizes variation across cases rather than uncertainty from repeated generations.

### Statistical analysis

Scores were displayed as means with one standard error of the mean (SEM). Both JAMA and PUMCH benchmarks used two-sided paired t-tests based on case-level scores. Each contrast paired the reference and control responses to the same evaluation case. For paired differences d, the test statistic was $t = \bar{d}/(s_d/\sqrt{n})$, where s_d is their sample standard deviation. JAMA tests used all 100 cases (df = 99), and PUMCH tests used all 123 cases (df = 122). The reference was KRAL for framework comparisons and full GRPO for component ablations.

Each benchmark included five framework contrasts and four no-RAG component-ablation contrasts without multiplicity adjustment; $P < 0.05$ was the nominal threshold. With-RAG component-ablation comparisons remained descriptive. SEM summarized variation across cases rather than uncertainty across training seeds.

## Results

### Model development and training data

Distillation yielded 37,117 question–answer records, of which 36,971 remained after question-level deduplication. These supplied 35,122 training and 1,849 validation records for warmup SFT without JAMA data. Filtering retained 696 of 710 teacher-generated inpatient trajectories for GRPO. The resulting training collections supported the SFT and GRPO configurations described in Tables 1 and 2.

## Framework performance on PUMCH antimicrobial tasks

KRAL achieved the highest overall score (0.895) on the PUMCH antimicrobial benchmark, compared with 0.862 for GRPO without inference-time retrieval, 0.848 for SFT+RAG, 0.804 for SFT, 0.819 for BASE+RAG, and 0.748 for BASE (Figure 2 and Table 3). The gains over GRPO, SFT+RAG, and SFT were 0.033, 0.047, and 0.091, respectively. Two-sided paired t-tests across all 123 examples (df = 122) favored KRAL over GRPO (P = 0.00064), SFT+RAG (P = 0.00055), SFT ($P = 1.1 \times 10^{-6}$), BASE+RAG ($P = 4.7 \times 10^{-5}$), and BASE ($P = 4.1 \times 10^{-10}$).

## Framework performance on public clinical vignettes

On JAMA, GRPO without inference-time retrieval achieved the highest score (0.629), followed by SFT (0.606) and KRAL (0.600). SFT+RAG, BASE, and BASE+RAG scored 0.571, 0.509, and 0.467, respectively (Figure 2 and Table 3).

In two-sided paired comparisons across 100 cases (df = 99), KRAL exceeded BASE+RAG (mean difference, 0.133; P = 0.0048) and BASE (mean difference, 0.091; P = 0.0499). Differences from GRPO, SFT+RAG, and SFT were not significant (P = 0.381, 0.315, and 0.861, respectively). JAMA cases were excluded from training and covered broader clinical tasks than the PUMCH-focused training data.

## Retrieval effects across benchmarks

Adding inference-time RAG increased PUMCH scores by 0.071 for BASE, 0.044 for SFT, and 0.033 for GRPO. On JAMA, the corresponding differences were −0.042, −0.035, and −0.029. These within-configuration differences were calculated from the framework scores in Table 3; no formal interaction test was performed.

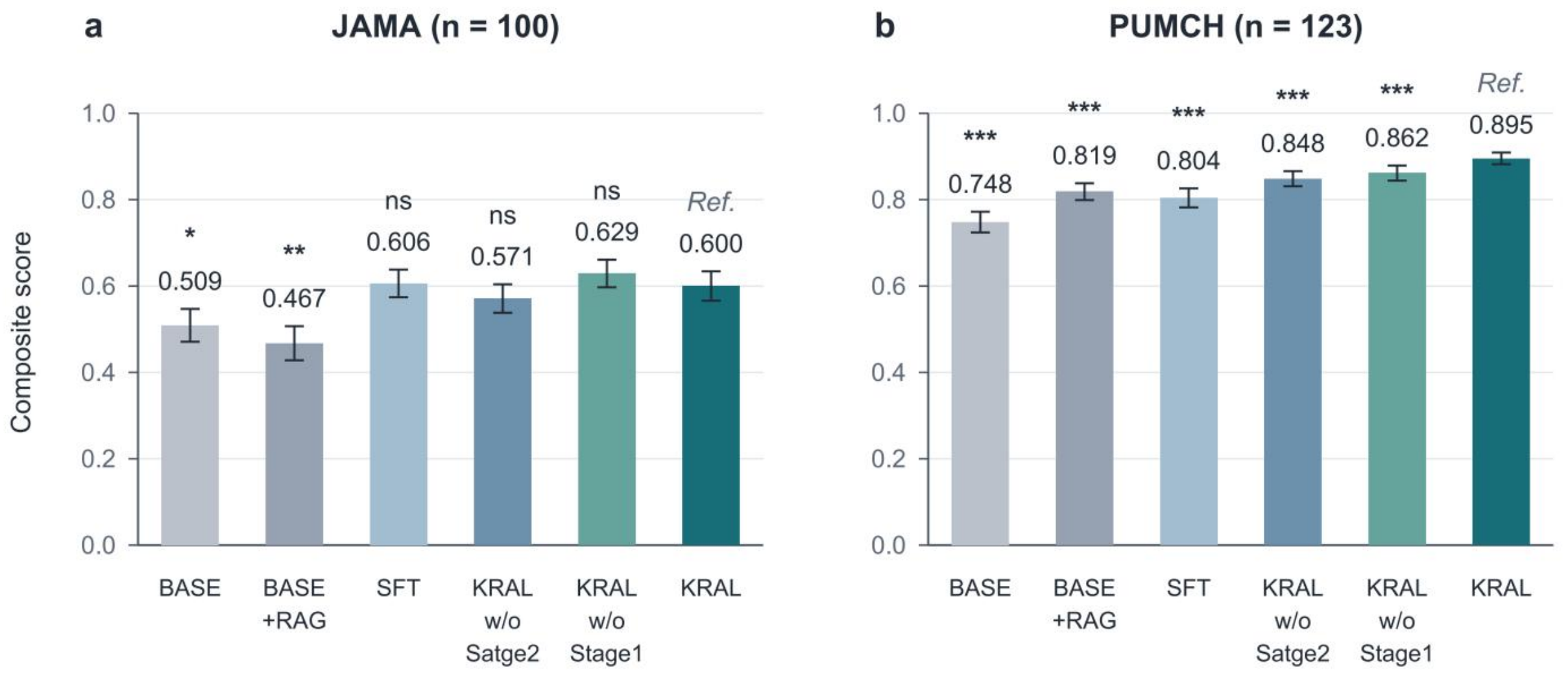

*Figure 2. Framework comparison on JAMA (a) and PUMCH (b). KRAL = GRPO+RAG; KRAL w/o Stage1 = GRPO; KRAL w/o Stage2 = SFT+RAG. Error bars show ±1 SEM across cases. Two-sided paired t-tests used 100 JAMA cases (df = 99) and 123 PUMCH examples (df = 122). *P < 0.05; **P < 0.01; ***P < 0.001; ns, P ≥ 0.05. All tests are unadjusted. All comparisons are against KRAL (Ref.).*

## GRPO component ablations

Without inference-time retrieval, full GRPO had higher composite scores than all four component ablations on both benchmarks (Figure 3 and Table 4). On JAMA, removing the higher clipping bound produced a score of 0.626 versus 0.627 for full GRPO; scores were 0.605 without subword matching, 0.583 without the repetition penalty, and 0.578 without reward smoothing. Corresponding PUMCH Antimicrobial scores were 0.790, 0.808, 0.830, and 0.814, compared with 0.841 for full GRPO. Exploratory paired contrasts against full GRPO on JAMA gave P = 0.969, 0.380, 0.104, and 0.022 for removal of higher clipping, subword Jaccard, the repetition penalty, and reward smoothing, respectively. The corresponding PUMCH Antimicrobial paired contrasts across 123 examples (df = 122) gave P = 0.0021, 0.056, 0.420, and 0.043. Thus, reward-smoothing removal met the nominal threshold on both benchmarks, and higher-clipping removal also did so on PUMCH Antimicrobial; the remaining no-RAG contrasts did not. With RAG, the full configuration remained highest at 0.875, while removal of the repetition penalty yielded a similar score of 0.872 (Table 4). The above ablation experiments support the effectiveness of the higher clipping bound and reward smoothing.

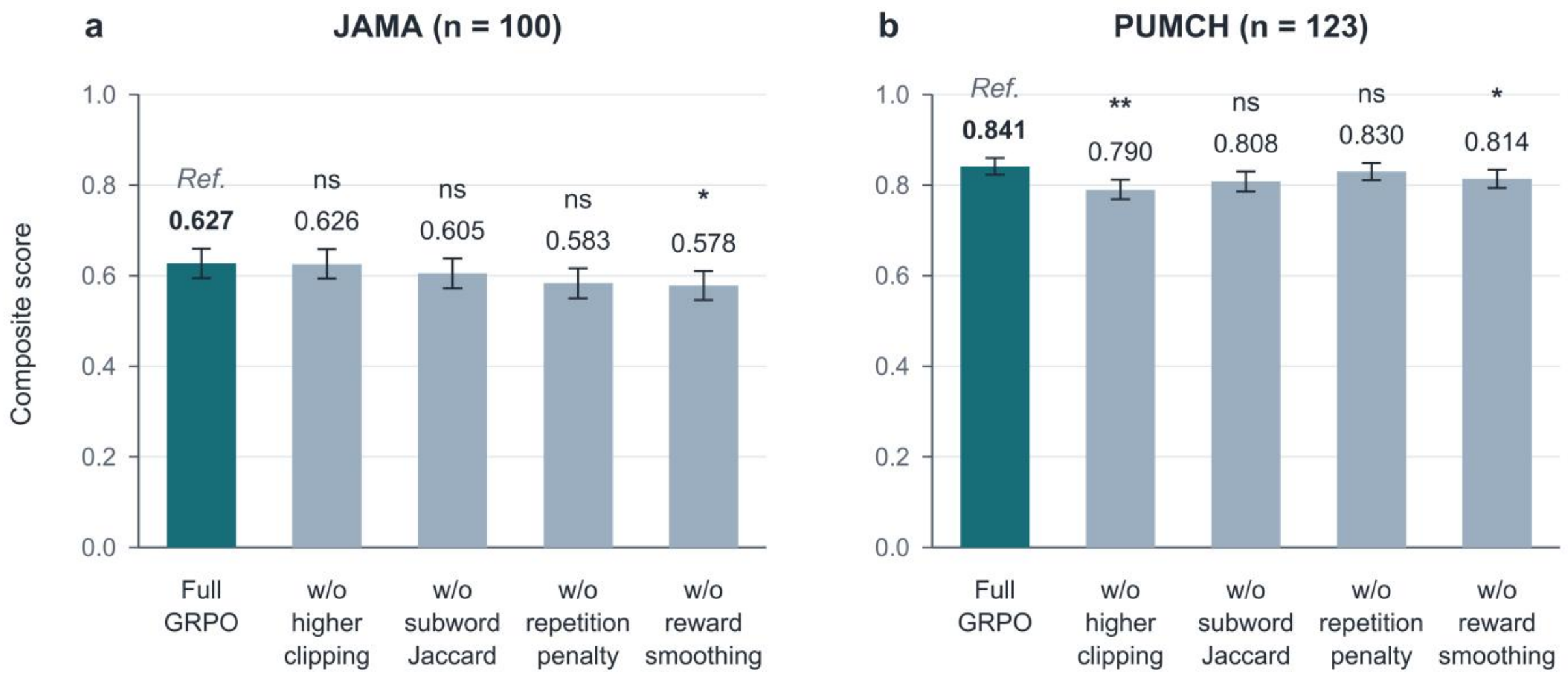

*Figure 3. GRPO component ablations on JAMA (a) and PUMCH (b). All conditions were evaluated without inference-time RAG. Error bars show ±1 SEM across cases. Labels compare each ablation with full GRPO (Ref.): *P < 0.05; **P < 0.01; ns, P ≥ 0.05. Tests were two-sided paired t-tests and unadjusted, with df = 99 for 100 JAMA cases and df = 122 for all 123 PUMCH cases. Higher clipping denotes the asymmetric upper clipping bound; subword denotes character/subword Jaccard matching.*

## GRPO training stability

Full GRPO remained stable through the 3,500-step budget (Figure 4). Approximate gradient-instability onset was 600 steps without reward smoothing, 1,700 without the higher clipping bound, 2,200 without subword matching, and 3,300 without the repetition penalty. Full GRPO achieved a failure-penalized reward AUC of 0.745 with 100% valid-budget coverage. The clipping, subword, repetition, and smoothing ablations achieved AUC/coverage values of 0.392/54%, 0.473/66%, 0.648/94%, and 0.034/18%, respectively (Table 4). Coverage summarized valid reward contributions across the budget, whereas instability onset described an approximate point in training.

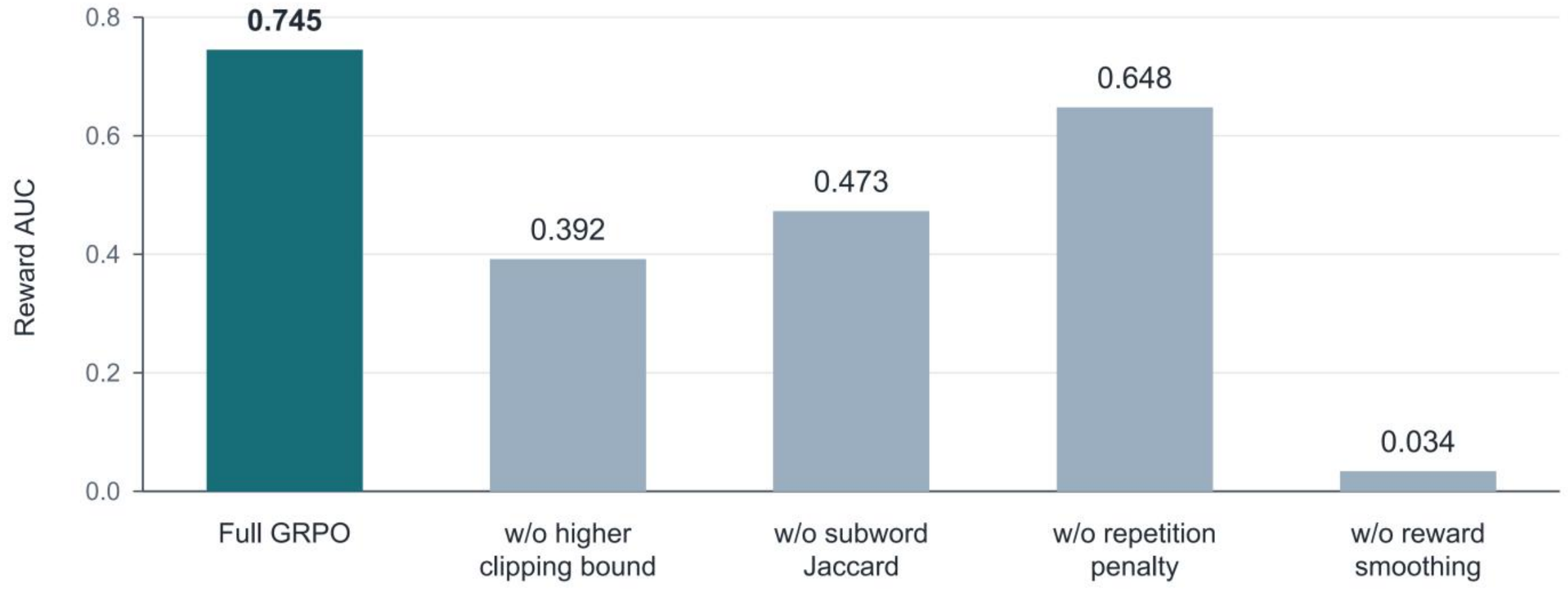

*Figure 4. Failure-penalized reward AUC for full GRPO and its component ablations. Budget: 3,500 steps. Invalid rewards and post-failure steps contribute zero. Valid-budget coverage was 100%, 54%, 66%, 94%, and 18% for full GRPO, without higher clipping, without subword Jaccard, without repetition penalty, and without reward smoothing, respectively.*

## Discussion

KRAL achieved the highest score on the PUMCH clinical benchmark and significantly outperformed all five framework controls. This finding is consistent with a benefit from combining guideline-grounded learning with case-based reasoning optimization. The comparison with SFT reflects the combined effects of GRPO and inference-time retrieval within the full framework. Component ablations further informed the training design: full GRPO completed the allocated budget, whereas all four ablations developed gradient instability, although the magnitude of benchmark-score differences varied.

Both the JAMA and PUMCH benchmark datasets were strictly isolated from the SFT and GRPO training data, and therefore served as external benchmarks. However, on the JAMA dataset, KRAL only outperformed the two base-model configurations but did not significantly outperform SFT or GRPO. A possible explanation is that the guidelines in the local knowledge base are those strictly followed in in-hospital clinical practice, and their recommendations may differ from the guideline basis underlying the reference answers in the JAMA dataset. Further work is needed to attribute the sources of these differences.

The framework comparisons address a different question from antimicrobial recommendation and guideline-chatbot evaluations [4–6]: whether model adaptation changes the usefulness of external evidence. Pairing each training level with and without distillation or GRPO isolate the independent contributions of Stage 1, Stage 2. The component ablations suggest that reward design and clipping affected whether training completed its allocated budget. Removing reward smoothing or the higher clipping bound substantially reduced failure-penalized reward AUC. Training reward, completion of the compute budget, and held-out task scores therefore describe different aspects of model behavior.

Clinical utility and safety require validation beyond the present benchmark comparisons. Automated judge scores do not adequately characterize contraindications, allergy or interaction conflicts, dose errors, route, treatment duration, or delayed escalation. Independent clinicians should assess these errors and their severity before prospective evaluation of clinician reliance and workflow effects. Prospective studies should also examine calibration and abstention, using guidance such as DECIDE-AI [26]. Local student deployment requires institutional governance, and teacher generation or judging through external services requires separate privacy review.

In conclusion, KRAL integrates guideline-grounded learning, warmup SFT, and Multi-turn Agentic GRPO for antimicrobial decision support. It outperformed all five controls on the PUMCH benchmark and sustained stable GRPO training. Independent clinical assessment and multicenter prospective evaluation are needed before clinical implementation.

## Funding

This work was supported by the Beijing Municipal Natural Science Foundation (L222019); CAMS Innovation Fund for Medical Sciences (CIFMS) (2024-I2M- C&T-C-002); National High Level Hospital Clinical Research Funding (2022-PUMCH-B-115&2022-PUMCH-D-005); and National Key R&D Program of China (2024YFF1207104).

## Acknowledgments

Group authorship: China Critical Care Clinical Trials Group (CCCCTG) and China National Critical Care Quality Control Center Group (China-NCCQC group).

## Data and code availability

JAMA Clinical Challenge source articles are available through JAMA Network [17], subject to publisher access and reuse conditions. Reproducible release should include the selected 100-case evaluation manifest, the separate 200-case JAMA collection manifest, and a case-level cross-set deduplication audit; the separate collection was not part of SFT. PUMCH sources and benchmark records remain access-restricted. A patient- and record-level audit linking the 710 source records and the 123 evaluation cases should be retained under institutional access controls. The experimental record identifies generation, scoring, training, and result files, as well as a KRAL-rollout Conda environment export and per-case output, judge, and summary JSON files. Public availability of those artifacts was not established. Reproduction requires the permitted code, manifests, model revisions, and paired predictions.

# Tables

Table 1. Data sources and experimental roles

| Source | Processing and retained data | Analysis role | Access |
|---|---|---|---|
| PUMCH guidance | 750 pages; 4,157 chunks; 4,153 usable chunks generated 12,459 questions. | Shared retrieval corpus and question generation | Institutional |
| PUMCH CDSS | 84 candidate questions; 79 retained; 12,538 seeds after merging with guideline questions. | Question expansion and distillation | Institutional |
| Distilled CoT | 37,251 questions → 37,117 answer records → 36,971 unique questions; 35,122 train / 1,849 validation. | SFT supervision | Derived institutional |
| PUMCH inpatient records | 2024 admissions with antimicrobial orders and infection-related diagnoses; one admission per patient; 710 records → 696 filtered trajectories. | GRPO reference actions and responses | Institutional |
| Separate JAMA collection | 200 unique cases; previous 180/20 designations; case-by-case deduplication against 100 evaluation cases. | Separate collection; excluded from SFT | Publisher conditions |
| JAMA evaluation | 100 selected English clinical vignettes; options removed. | Diagnosis or next-step evaluation | Publisher conditions |
| PUMCH evaluation | 123 clinician-validated clinical antimicrobial treatment cases; no patient- or record-level overlap with the 710 source records. | Clinical benchmark evaluation | Institutional |

*PUMCH: Peking Union Medical College Hospital; CDSS: clinical decision support system; CoT: chain of thought; SFT: supervised fine-tuning; GRPO: Group Relative Policy Optimization.*

Table 2. Supervised and reinforcement learning configurations

| Setting | Warmup SFT | GRPO |
|---|---|---|
| Input records | 35,122 CoT; 1,849 held-out validation; no JAMA | 696 trajectories |
| Initialization | Qwen3.8-27B | SFT checkpoint |
| LoRA rank / alpha | 64 / 128 | 16 / 32 |
| Learning rate | $2 \times 10^{-4}$ | $5 \times 10^{-6}$ |
| Warmup fraction | 0.05 | 0.15 |
| GPU allocation | 8 training | 4 training + 4 rollout |
| DeepSpeed / precision | ZeRO-3 / bfloat16 | ZeRO-2 / bfloat16 |
| Microbatch / accumulation | 2 / 4 | 1 / 2 |
| Gradient clipping | 1.0 | 0.3 |
| Length limit | 8,192 sequence tokens | 20,480 total; 4,096 new |
| Planned duration | 3 epochs | 3,500 steps; ≤3 epochs |
| Observed duration | Not reported for the corrected dataset | 3,500 steps for full model |
| Validation / selection | 1,849 held-out CoT records; selection step not specified | Best 20-step mean reward |
| Other settings | Dropout 0.05; weight decay 0.1; NEFTune 5 | 8 responses/case; ≤4 retrieval rounds; KL 0.001; clipping 0.1/0.4 |

*SFT: supervised fine-tuning; GRPO: Group Relative Policy Optimization; CoT: chain of thought; LoRA: low-rank adaptation; KL: Kullback–Leibler divergence. Platform: HGX H200, 8 × H200 SXM (~141 GB/card), 2.0 TiB RAM, Ubuntu 22.04.5. Reported environment: driver 595.91.07; CUDA driver capability 13.2, Toolkit 12.4.131; Python 3.11.15; PyTorch 2.11.0; vLLM 0.23.0; Transformers 5.16.1; Accelerate 1.14.0; DeepSpeed 0.16.9; TRL 0.29.1; PEFT 0.19.1; FlashAttention 2.7.4.post1.*

Table 3. Framework comparison on public and PUMCH evaluations

| Condition | JAMA mean ± SEM (n = 100) | PUMCH composite mean ± SEM (n = 123) |
|---|---|---|
| BASE | 0.509 ± 0.038 | 0.748 ± 0.024 |
| BASE+RAG | 0.467 ± 0.040 | 0.819 ± 0.019 |
| SFT | 0.606 ± 0.032 | 0.804 ± 0.022 |
| SFT+RAG (KRAL w/o Stage2) | 0.571 ± 0.033 | 0.848 ± 0.017 |
| GRPO (KRAL w/o Stage1) | 0.629 ± 0.032 | 0.862 ± 0.018 |
| KRAL (full GRPO+RAG) | 0.600 ± 0.034 | 0.895 ± 0.014 |

*All composite scores are on a 0–1 scale and reflect graded response quality rather than proportions of correct answers. Values are mean ± SEM across cases; PUMCH means were calculated across all 123 cases. RAG: retrieval-augmented generation; SFT: supervised fine-tuning; GRPO: Group Relative Policy Optimization; KRAL: Knowledge and Reasoning Augmented Learning. KRAL = full GRPO+RAG; SFT+RAG is labeled KRAL w/o Stage2 in Figure 2, and GRPO without inference-time RAG is labeled KRAL w/o Stage1. All GRPO configurations retain warmup SFT initialization. Two-sided paired tests versus KRAL used df = 99 for JAMA and df = 122 for PUMCH. In table order before KRAL, JAMA P values are 0.0499, 0.0048, 0.861, 0.315, and 0.381; PUMCH P values are $4.1 \times 10^{-10}$, $4.7 \times 10^{-5}$, $1.1 \times 10^{-6}$, 0.00055, and 0.00064. All tests are unadjusted.*

Table 4. GRPO component ablations and training stability

| Configuration | JAMA no RAG | PUMCH no RAG | PUMCH with RAG | Reward AUC | Valid budget |
|---|---|---|---|---|---|
| Full GRPO | 0.627 | 0.841 | 0.875 | 0.745 | 100% |
| Without higher clipping bound | 0.626 | 0.790 | 0.846 | 0.392 | 54% |
| Without subword Jaccard | 0.605 | 0.808 | 0.845 | 0.473 | 66% |
| Without repetition penalty | 0.583 | 0.830 | 0.872 | 0.648 | 94% |
| Without reward smoothing | 0.578 | 0.814 | 0.847 | 0.034 | 18% |

*GRPO: Group Relative Policy Optimization; RAG: retrieval-augmented generation; PUMCH: Peking Union Medical College Hospital; AUC: area under the curve. Benchmark sizes: JAMA 100; PUMCH 123. Reward AUC is the failure-penalized mean over a fixed 3,500-step budget, assigning invalid and post-failure steps zero reward. Valid budget is the proportion of steps contributing valid rewards. All variants used the best available reward checkpoint. Benchmark scores and reward AUC use different definitions and are not interchangeable; no repeated-seed uncertainty is implied. Figure 3 adds SEM error bars and exploratory no-RAG contrasts against full GRPO. In table order after full GRPO, JAMA SEMs are 0.032, 0.033, 0.033, and 0.032 (reference SEM, 0.032), with $P = 0.969$, 0.380, 0.104, and 0.022. PUMCH SEMs are 0.022, 0.022, 0.019, and 0.020 (reference SEM, 0.019), with $P = 0.0021$, 0.056, 0.420, and 0.043. JAMA and PUMCH tests were two-sided paired t-tests with $df = 99$ and $df = 122$, respectively. All P values are unadjusted; with-RAG scores remain descriptive.*